%% file: main.tex
\documentclass[10pt,twocolumn,letterpaper]{article}
\usepackage{iccv}      
\input{preamble}
\definecolor{iccvblue}{rgb}{0.21,0.49,0.74}
\usepackage[pagebackref,breaklinks,colorlinks,allcolors=iccvblue]{hyperref}

\def\confName{ICCV}
\def\confYear{2025}

\title{Auto-Vocabulary Semantic Segmentation} 
\author{Osman Ülger\textsuperscript{1}\thanks{Equal contribution.} \quad Maksymilian Kulicki\textsuperscript{2}\footnotemark[1] \quad Yuki Asano\textsuperscript{3} \quad Martin R. Oswald\textsuperscript{1}\\
{\small \textsuperscript{1} University of Amsterdam} \quad
{\small\textsuperscript{2} Polish Academy of Sciences} \quad
{\small\textsuperscript{3} University of Technology Nuremberg}}

\begin{document}
\twocolumn[{%
  \maketitle
  \vspace{-15pt}
  \captionsetup{type=figure}
  \centering
  \footnotesize
  \setlength{\tabcolsep}{1pt}
  \renewcommand{\arraystretch}{1}
  \newcommand{\sz}{0.235}
  \newcommand{\hz}{2.4cm}
  \begin{tabular}{rcc@{\hskip 5pt}cc}
    \raisebox{12pt}{\rotatebox{90}{\makecell{Fixed-Set\\ Ground Truth}}} &
    \includegraphics[width=\sz\linewidth,height=\hz]{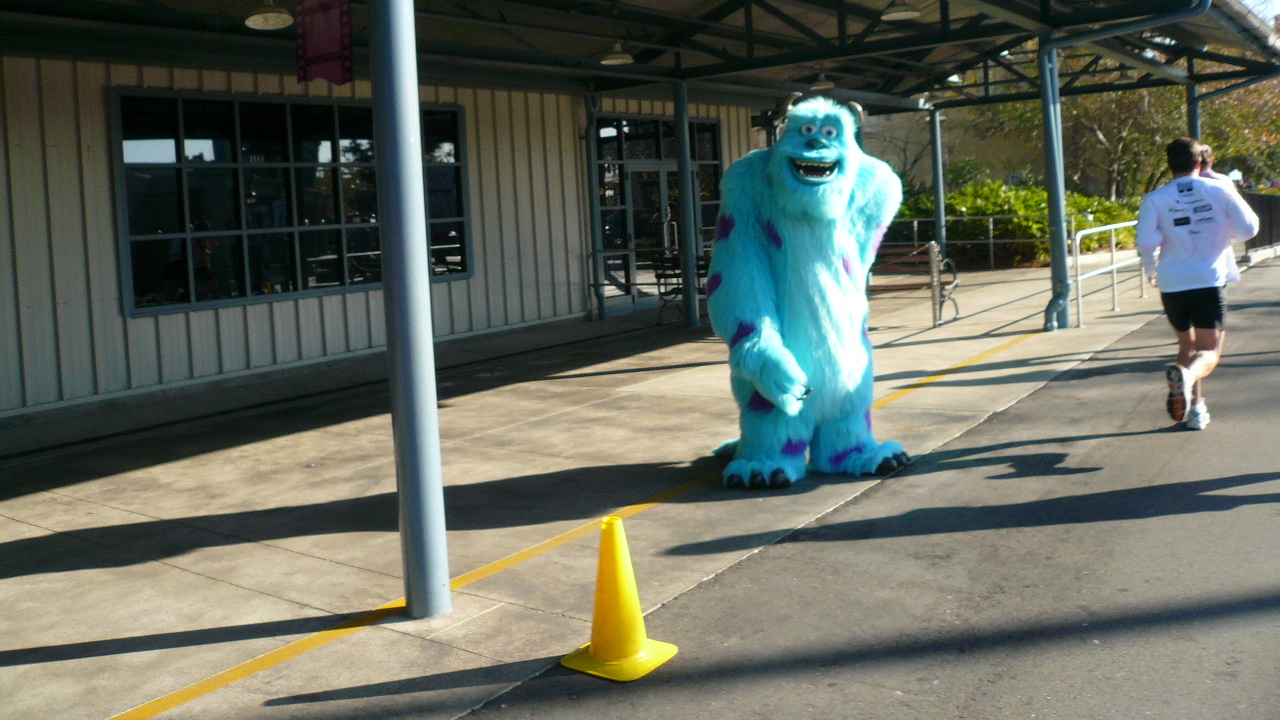} &
    \includegraphics[width=\sz\linewidth,height=\hz]{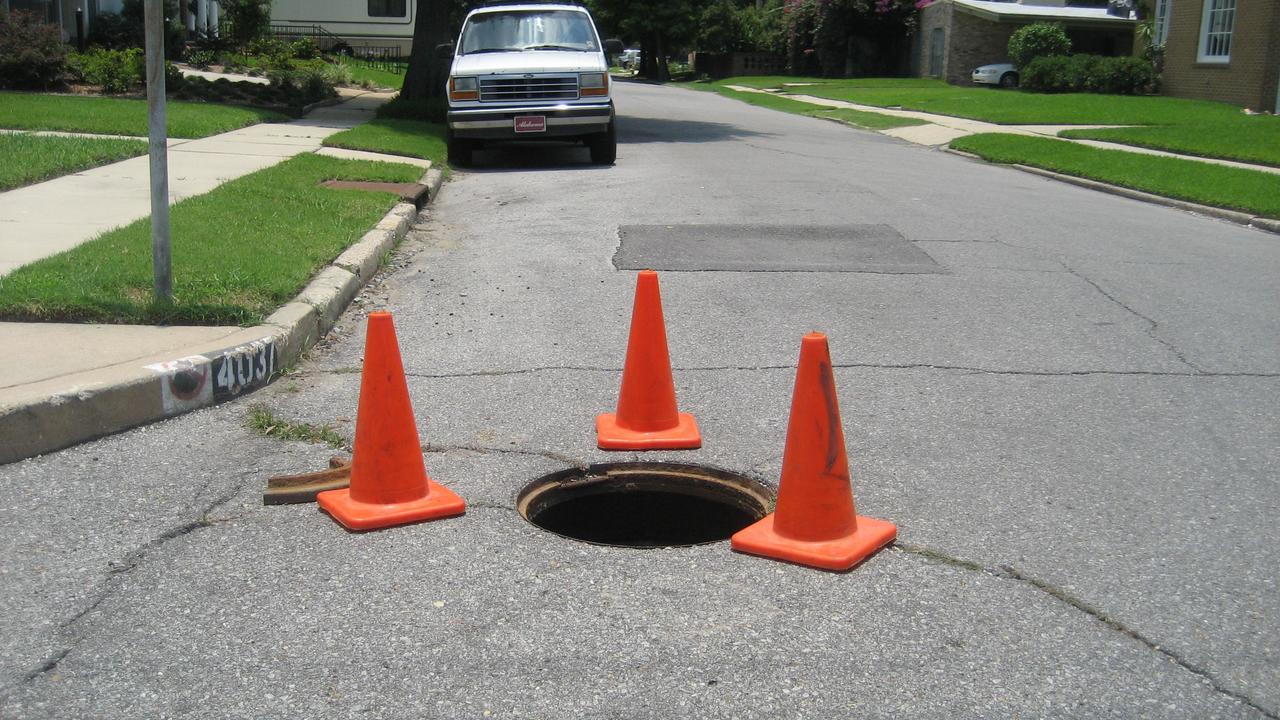} &
    \includegraphics[width=\sz\linewidth,height=\hz]{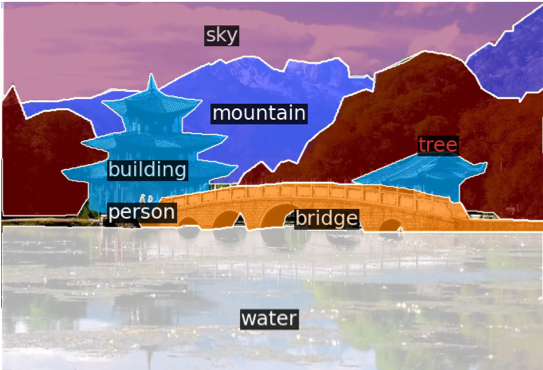} &
    \includegraphics[width=\sz\linewidth,height=\hz]{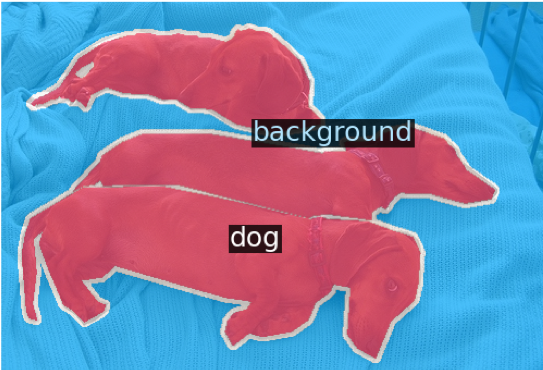} \\[3pt]
    \raisebox{10pt}{\rotatebox{90}{\makecell{AutoSeg (Ours)}}} &
    \includegraphics[width=\sz\linewidth,height=\hz]{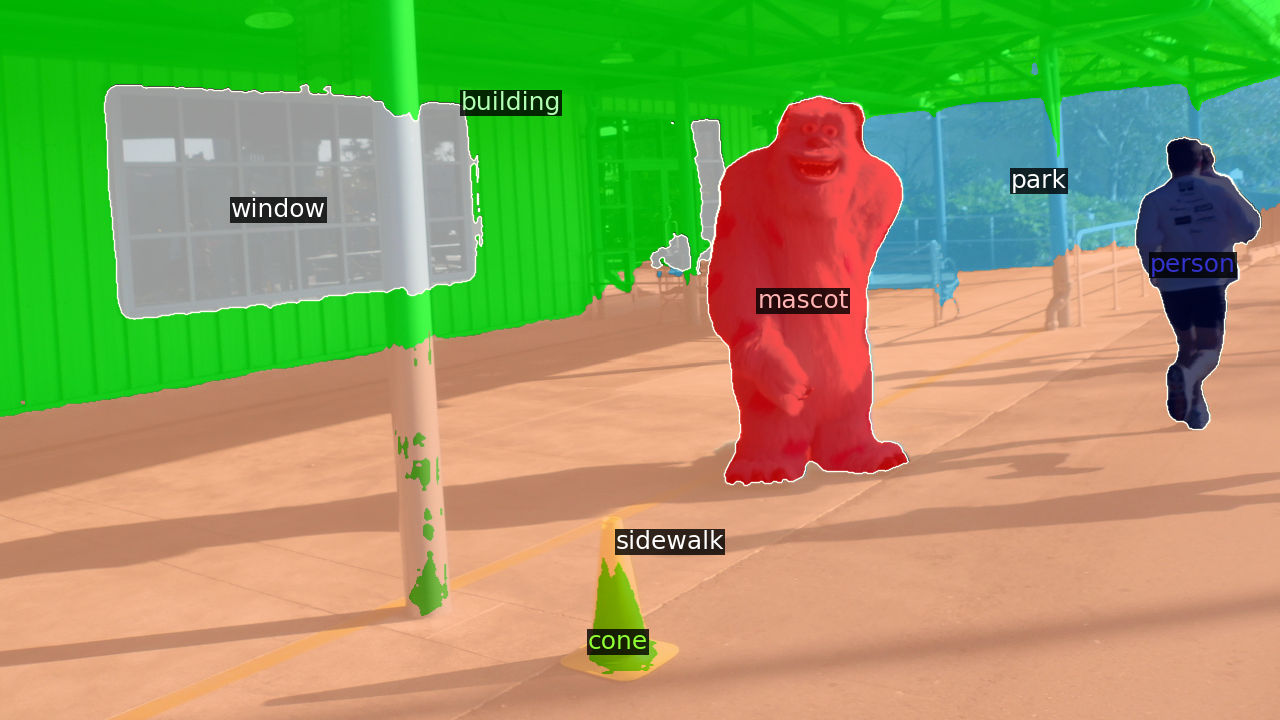} &
    \includegraphics[width=\sz\linewidth,height=\hz]{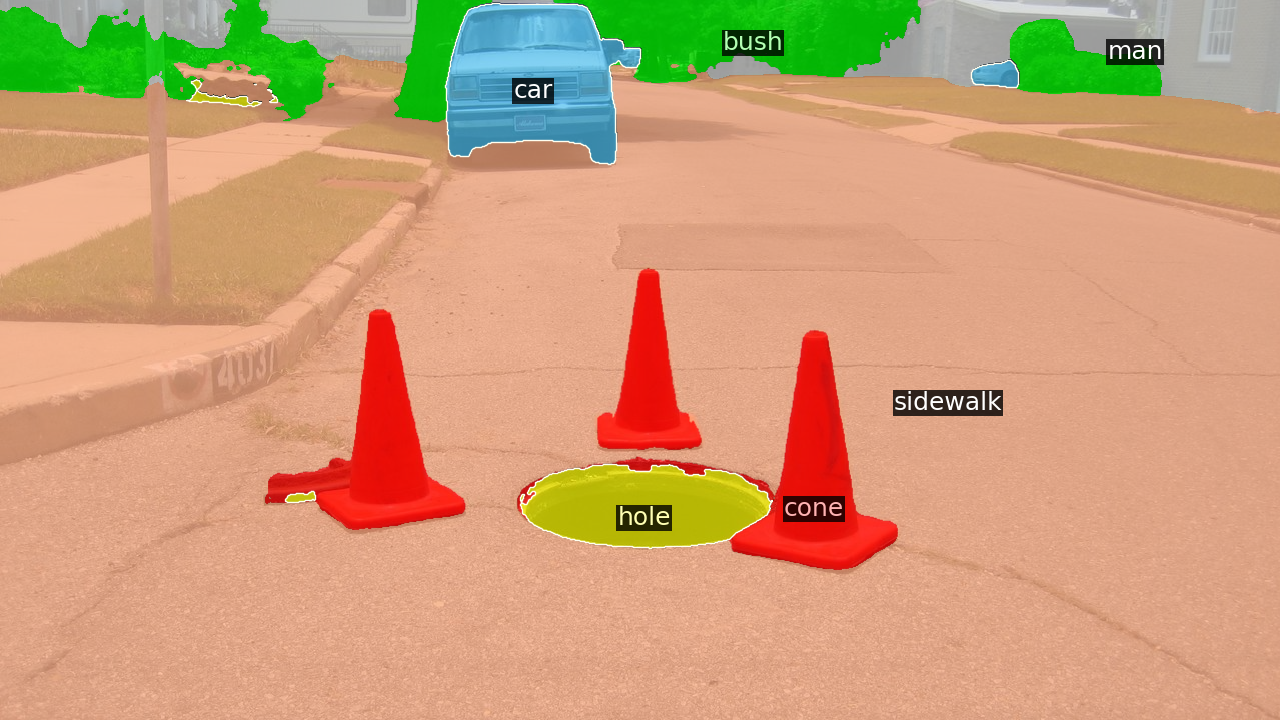} &
    \includegraphics[width=\sz\linewidth,height=\hz]{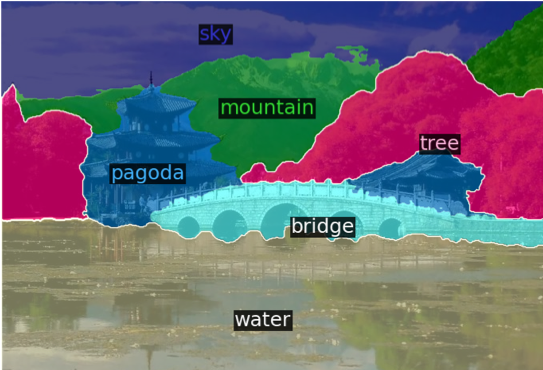} &
    \includegraphics[width=\sz\linewidth,height=\hz]{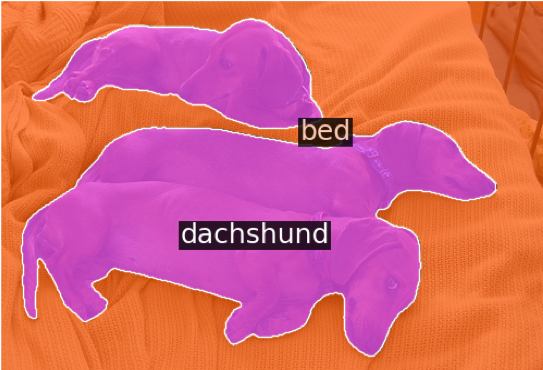} \\[-1pt]
    & \multicolumn{2}{c}{Open-Ended Predictions} & \multicolumn{2}{c}{Out-of-Vocabulary Predictions}\\[-2pt]
  \end{tabular}
  \captionof{figure}{\textbf{\ours Exemplary Results.} AutoSeg is readily applicable to unseen images for open-ended segmentation for objects such as \textit{mascot} and \textit{hole}, such as the two images on the left. Furthermore, where established segmentation datasets have a fixed set of annotation categories, our method is able to identify and segment with more semantically precise object categories beyond the fixed-set ground truth, such as \textit{dachshund}, \textit{bed} and \textit{pagoda}.
  Images are from the Road Anomaly~\cite{roadanomaly}, PASCAL~\cite{VOC} and ADE20K~\cite{ADE20k} datasets.}
  \label{fig:teaser}
  \vspace{2em}
}]

\maketitle
\input{Sections/0_abstract}
\input{Sections/1_introduction}
\input{Sections/2_related_work}
\input{Sections/3_method}
\input{Sections/4_experiments}
\input{Sections/5_conclusion}
\input{Sections/6_acknowledgements}

{
    \small
    \bibliographystyle{ieeenat_fullname}
    \bibliography{main}
}

\clearpage
\input{Sections/x_supplementary}    

\end{document}

%% file: preamble.tex
\usepackage{graphicx}
\usepackage{amsmath}
\usepackage{amssymb}
\usepackage{booktabs}
\usepackage{makecell}
\usepackage{colortbl}
\usepackage{bbm}
\usepackage{float}
\usepackage{pifont}
\usepackage{algorithm}
\usepackage{algorithmic}
\usepackage{multirow}
\usepackage{bm}
\usepackage{listings}

\newcommand{\xmark}{\ding{55}}%
\newcommand{\noo}{\textcolor{red}{\xmark}}
\newcommand{\yes}{\textcolor{OliveGreen}{\checkmark}}

\def\eg{\emph{e.g.}\ } 
\def\ie{\emph{i.e.}\ }

\newcommand{\ours}{AutoSeg\xspace}
\newcommand{\blip}{BBoost\xspace}

\colorlet{colorFst}{Green!25}       
\colorlet{colorSnd}{SpringGreen!45} 
\colorlet{colorTrd}{Yellow!30}      
\colorlet{colorLow}{darkgray!30}    

\newcommand{\fs}{\cellcolor{colorFst}\bf}   
\newcommand{\nd}{\cellcolor{colorSnd}}      
\newcommand{\rd}{\cellcolor{colorTrd}}      

\newcommand{\boldparagraph}[1]{\vspace{0.1em}\noindent{\bf #1}}

%% file: Sections/0_abstract.tex
\begin{abstract}
Open-Vocabulary Segmentation (OVS) methods are capable of performing semantic segmentation without relying on a fixed vocabulary, and in some cases, without training or fine-tuning. However, OVS methods typically require a human in the loop to specify the vocabulary based on the task or dataset at hand. In this paper, we introduce Auto-Vocabulary Semantic Segmentation (AVS), advancing open-ended image understanding by eliminating the necessity to predefine object categories for segmentation. Our approach, \ours, presents a framework that autonomously identifies relevant class names using semantically enhanced BLIP embeddings and segments them afterwards. Given that open-ended object category predictions cannot be directly compared with a fixed ground truth, we develop a Large Language Model-based Auto-Vocabulary Evaluator (LAVE) to efficiently evaluate the automatically generated classes and their corresponding segments. With AVS, our method sets new benchmarks on datasets PASCAL VOC, Context, ADE20K, and Cityscapes, while showing competitive performance to OVS methods that require specified class names. All code will be publicly released.
\end{abstract}
\footnotetext[1]{Equal contribution.}

%% file: Sections/1_introduction.tex
\section{Introduction}
While humans possess an open-ended understanding of scenes, recognizing thousands of distinct categories, semantic segmentation methods~\cite{minaee2021image} typically rely on a fixed vocabulary of predefined semantic categories. They require large human-annotated datasets and have limited capabilities for handling a broad range of classes or unknown objects. Recent studies have focused on addressing these limitations~\cite{weak-sup1, self-sup1} with models that leverage Vision-Language Models (VLMs) as an emerging category~\cite{lseg, openseg, maskclip, ovseg, catseg, fcclip, opsnet,san, simseg}. Such VLMs learn rich multi-modal features from large numbers of image-text pairs.
\begin{figure*}[ht!]
  \centering
  \includegraphics[width=\linewidth]{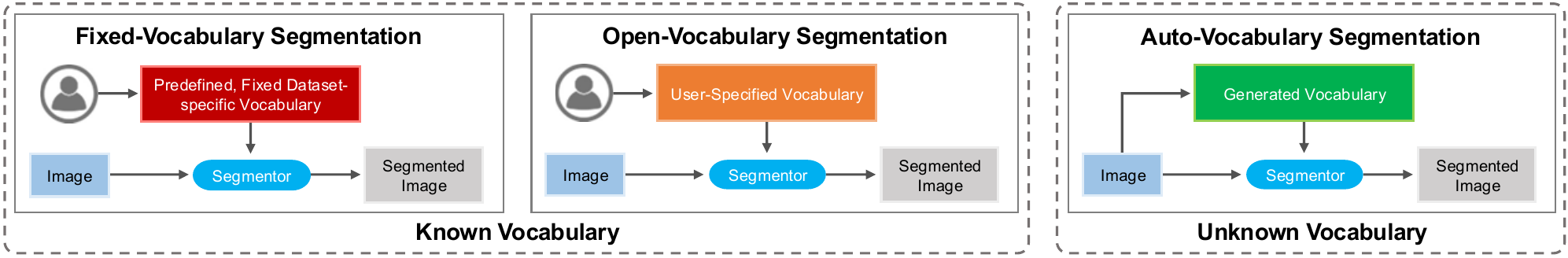}\\
  \caption{\textbf{Semantic Segmentation Tasks in Comparison.} In traditional Semantic Segmentation, an image is segmented into fixed, predefined set of classes (fixed vocabulary). In Open-Vocabulary Segmentation, the user specifies which object categories (from the open vocabulary) should be segmented: 1) either via a human-provided prompt at runtime, or 2) the OV-method is trained to output the vocabulary of a human-annotated target dataset. In contrast, Auto-Vocabulary Segmentation automatically generates relevant object categories directly from the image. This enables true open-ended scene understanding without needing human input.}
  \label{fig:diff_seg_tasks}
  \vspace{-5pt}
\end{figure*}
Due to the immense computational training costs of VLMs, it is common to build upon pre-trained VLMs such as CLIP~\cite{clip} or BLIP~\cite{blip}. However, applying these VLMs on per-pixel tasks to obtain precise locality information is non-trivial, as they are trained on full images and lack the ability to directly reason over local regions in the image. This makes them less useful for obtaining precise segmentation boundaries, which is crucial for downstream tasks like robot grasping. Moreover, current Open-Vocabulary segmentation (OVS) methods still require a human in the loop, inherently limiting the scalability of these methods (see Fig.~\ref{fig:diff_seg_tasks}). OVS methods can be separated into two categories: \textbf{1) prompt-based methods} (\eg \cite{SAM,lisa,xdecoder}) require per-image user input to provide a \emph{known} vocabulary; and \textbf{2) dataset-based methods} (\eg \cite{openseg,ovseg,lseg,odise}) require human-annotated datasets in combination with training in which the \emph{known} vocabulary is baked into the method as its defined output categorization. For both categories, a human is providing a \emph{known} and fixed vocabulary, which drastically simplifies the segmentation problem as the ground truth vocabulary is always provided, but it also limits the application use-cases when a human needs to be in the loop. Imagine a kitchen robot that needs to distinguish a large variety of tools and ingredients when cooking a recipe and requiring precise segmentations for grasping. It is undesirable to require a human-provided prompt for every grasp or to re-label a dataset and re-train the segmentation method every time a new recipe contains a new tool or ingredient.

To eliminate the human in the loop and fully rely on pre-trained foundation models, we propose \textbf{Auto-Vocabulary Semantic Segmentation (AVS)}. In AVS, the \emph{unknown} vocabulary is automatically generated and integrated into a semantic segmentation method thus allowing pixel-level classifications for \textbf{any} class without the need for textual input from the user, predefined class names, additional data, training or fine-tuning. To address this task, we introduce \ours, a zero-shot method that first identifies image-specific target categories and then predicts masks for them.
See Fig.~\ref{fig:teaser} for exemplary results.
We introduce \blip for vocabulary generation, which provides a simple yet effective semantic clustering strategy to enhance locality and semantic precision when captioning with BLIP.

Furthermore, we address the performance evaluation of open-ended segmentation models, following the same goal of eliminating human input for targeting the true scalability of segmentation models.
While VLMs have a \emph{continuous} latent representation of semantics, the inherent nature of segmentation is to \emph{discretize} semantic categories to a particular task-specific vocabulary.
Comparing different discrete category sets is challenging, especially for larger vocabularies as exact matching can be difficult due to annotation ambiguities like synonyms (\eg table$\leftrightarrow$desk), semantic hierarchies (\eg person$\rightarrow$man$\rightarrow$groom), or spatial hierarchies (\eg car$\rightarrow$wheel$\rightarrow$rim).
Human-annotated datasets typically have an annotation bias towards these ambiguities and can be inconsistently annotated across multiple annotators, making it nearly impossible for a segmentation algorithm to infer the intended hierarchy level. The classical discrete semantic evaluation also neglects semantic vicinities among different labels, treating such ambiguities equal to misclassifications.
Since our methods' output is not within a pre-defined, but an auto-generated vocabulary, we enable evaluations on labeled datasets with our proposed \textbf{L}arge Language Model-based \textbf{A}uto-\textbf{V}ocabulary \textbf{E}valuator (\textsc{LAVE}) which maps a generated vocabulary to a provided one.

In summary, our \textbf{contributions} are as follows: 
\textbf{1)} we introduce \ours, a novel framework to automatically determine and segment open-ended classes in an image; \textbf{2)} we propose \blip, a novel method for generating image-specific target vocabularies, leveraging text decoding from enhanced BLIP embeddings; and \textbf{3)} we propose \textsc{LAVE}, a novel evaluation approach for auto-vocabulary semantic segmentation which utilizes an LLM. Through qualitative and quantitative analyses, we demonstrate the effectiveness of our framework for AVS on multiple public datasets.

%% file: Sections/2_related_work.tex
\begin{figure*}[t!]
  \centering
  \includegraphics[width=\linewidth]{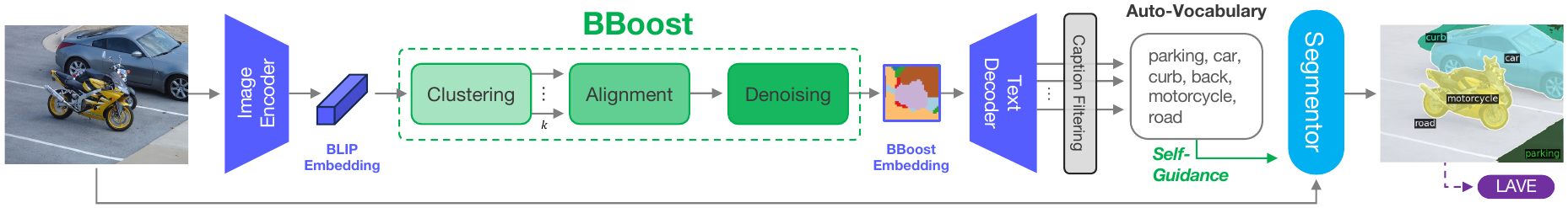}\\
  \caption{\textbf{Method Overview.} BLIP encodings are clustered, aligned and denoised before being decoded into nouns by \blip. Generated nouns serve as self-guidance to a segmentor, which predicts the final mask. When evaluating (purple), our custom evaluator LAVE processes the output, mapping predicted nouns to the fixed-vocabulary annotations.}
  \label{fig:pipeline}
  \vspace{-6pt}
\end{figure*}
\section{Related Work} \label{relatedwork}
\boldparagraph{Open-Vocabulary Segmentation (OVS).} OVS tackles the challenging task of segmenting images based on arbitrary text queries rather than a predefined class set. Early OVS methods, like ZS3Net~\cite{ZS3NET} and SPNet~\cite{SPNet}, treated it as a zero-shot task by training modules that align visual and linguistic embeddings, segmenting by comparing word vectors with local image features. Recently, Vision-Language Models (VLMs) pre-trained on large image-text dataset, such as Contrastive Language-Image Pre-training (CLIP)\cite{clip}, have unified image and text features. CLIP has advanced the field of OVS significantly, with methods such as LSeg\cite{lseg}, which compares per-pixel embeddings with class embeddings, and two-stage approaches like OpenSeg~\cite{openseg}, OPSNet~\cite{opsnet}, OVSeg~\cite{ovseg}, ZSSeg~\cite{zsseg}, and POMP~\cite{pomp}, which first obtain class-agnostic masks and then classify each mask with CLIP. Other works, such as MaskCLIP~\cite{maskclip}, FC-CLIP~\cite{fcclip}, and SAN~\cite{san}, utilize intermediate CLIP representations. Other approaches leverage CLIP to relate image semantics to class labels~\cite{catseg}, predict binary masks~\cite{simseg} or ensure segmentation consistency at multiple granularities~\cite{vlpart}. Beyond VLMs, recent models use text-to-image diffusion~\cite{ovdiff, odise} or transformers~\cite{groupvit, vilseg, ovsegmentor, xdecoder}, with X-Decoder providing a generalized model for various vision-language tasks.One common and significant limitation of the described methods is the need for textual input as a form of guidance by the user. Our work presents a novel framework which enables self-guidance by automatically identifying relevant object categories via localized image captioning. Closest and concurrent to our work is Zero-Guidance Segmentation~\cite{zeroguidance}, in which clustered DINO~\cite{dino} embeddings are combined with CLIP to achieve zero guidance by the user. Despite the similar setting, there are key differences: we use BLIP~\cite{blip} for both image clustering and caption generation, while they use a complex pipeline involving DINO clustering, CLIP embeddings, a custom attention-masking module, and CLIP-guided GPT-2 for captioning, requiring conversions between three latent representations. In contrast, we achieve superior segmentation quality with self-guidance using any out-of-the-box OVS model.\\
\boldparagraph{Image Captioning with Image-Text Embeddings.} Image Captioning is the task of describing the content of an image using natural language. A number of approaches have been proposed for the task, including ones utilizing VLMs. Approaches such as CLIP-Cap~\cite{clipcap} and CLIP-S~\cite{clips} utilize CLIP embeddings to guide text generation. More recently, the Large Language and Vision Assistant (LLaVA) was introduced with similar capabilities by connecting the visual encoder of CLIP with a language decoder, as well as training on multimodal instruction-following data~\cite{llava}. Other approaches rely on BLIP~\cite{blip}, a VLM in which the task of decoding text from image features was one of the tasks of the pretraining process. This enables BLIP to perform image captioning without additional components. Furthermore, the text decoder is guided by local patch embeddings in a way that enables local captioning based on a specific area. This unique capability, discovered during the development of our method, emerges organically in BLIP despite being trained solely with image-level captions. 

\boldparagraph{Visual Phrase Grounding.} Visual phrase grounding aims to connect different entities mentioned in a caption to corresponding image regions~\cite{rohrbach2016grounding}. This task resembles OVS, as it aims to find correspondences between text and image regions. However, predicted areas do not have to precisely outline object boundaries and can be overlapping. Visual grounding has been approached in a self-guided manner, in which heatmaps of regions of the image are generated from image-level CLIP embeddings, which are then captioned by the BLIP encoder~\cite{shaharabany2022looking}. One disadvantage of this approach is that the region determined by CLIP can indicate multiple objects at once, hence multiple objects are captioned for that region, preventing pixel-level class-specific predictions. In our work, BLIP serves both as encoder and decoder, enabling category-specific regions and captioning.

%% file: Sections/3_method.tex
\section{Method}
This work aims to perform semantic segmentation without any additional data, training, finetuning or pre-defined target categories. To this end, we propose to identify relevant object categories in an image by captioning with locality (Sect.~\ref{captioning}), filter the generated captions into a meaningful vocabulary 
and use the vocabulary as target categories for a segmentor (Sect.~\ref{selfguidancesect}). An overview is depicted in Fig.~\ref{fig:pipeline}.

\subsection{Local Region Captioning} \label{captioning}
To identify relevant object categories in the image, we employ Bootstrapped Language Image Pretraining (BLIP)~\cite{blip}. BLIP is a powerful VLM capable of performing accurate and detailed image captioning. Its image encoder is designed as a Vision Transformer~\cite{vit}, dividing the image into patches and encoding them into embeddings in which image-text features reside. The set of embeddings is then passed through transformer layers to capture contextual information across the image. Finally, all contextualized embeddings are decoded into text descriptions. Though effective in describing salient objects in the image, captioning models like BLIP generally fail to describe all elements of a scene comprehensively. This is likely due to the captions in the training data, which mostly focus on salient foreground objects. In turn, this limits its direct application to downstream tasks such as open-ended recognition, since many non-salient objects are missed. As a solution to this limitation, we propose \blip for exhaustive, local captioning. \blip clusters the encoded BLIP features into semantically meaningful feature clusters, which are enhanced to capture the object-specific feature representations more effectively. Afterwards, each individual feature cluster is fed to a pre-trained BLIP text decoder. This enables generating a description of each semantically distinct and meaningful area in the image, resulting in a more comprehensive, specific and accurate description of the image overall. Its components are detailed in the next paragraphs.

\boldparagraph{Clustering.} By default, BLIP expects an RGB image $\mathbf{X}_D \in \mathbb{R}^{384 \times 384 \times 3}$. Since common datasets often have images with higher resolution, we additionally process the image at a higher resolution $\mathbf{X}_H \in \mathbb{R}^{512 \times 512 \times 3}$. The multi-resolution set of images $\mathbf{X} = \{\mathbf{X}_D, \mathbf{X}_H\}$ is fed to the BLIP encoder to obtain the set of BLIP patch embeddings $\hat{\textbf{B}} = \{\hat{\textbf{B}}^{\mathbf{X}_D}, \hat{\textbf{B}}^{\mathbf{X}_H}\}$ at the two resolutions:
\begin{align}    
  \mathbf{P}^{\mathbf{X}_D} &= \left\{ \mathbf{X}^{D}_{ij} \,|\, \mathbf{X}^{D}_{ij} \in \mathbb{R}^{16 \times 16 \times 3}, 1 \leq i, j \leq 24 \right\} \\
  \mathbf{P}^{\mathbf{X}_H} &= \left\{ \mathbf{X}^{H}_{ij} \,|\, \mathbf{X}^{H}_{ij} \in \mathbb{R}^{16 \times 16 \times 3}, 1 \leq i, j \leq 32 \right\} \\
  \hat{\mathbf{B}}_{n}^{\mathbf{X}_R} &= T\left(f_{\text{MLP}}\left(\mathbf{P}_{n}^{\mathbf{X}_R}\right)\right) \parallel \mathbf{z}_{n}^{\mathbf{X}_R}, \notag \\
  &\quad \forall R \in \{D, H\}, \quad n \in \{1, \ldots, N_R\} \label{eq:sincos}
\end{align}
where $\textbf{P}^{\textbf{X}_D}\in \mathbb{R}^{576 \times 16 \times 16 \times 3}$ and $\textbf{P}^{\textbf{X}_H}\in \mathbb{R}^{1024\times 16 \times 16 \times 3}$ denote the sets of patches for resolutions $D$ 
and $H$ 
respectively, and $\mathbf{P}_{n}^{\mathbf{X}_R}$ is the $n$-th patch. $f_{\text{MLP}}(\cdot)$ represents a shared fully connected Multi-Layer Perceptron (MLP), $\textsc{LN}$ is LayerNorm, and $T$ is a Transformer encoder with $L$ alternating layers of Multi-Head Self-Attention (MHA) and an MLP, sequentially propagated:
\begin{align}
  T_l(X) &= X + \textsc{MHA}\big(\textsc{LN}(X), \textsc{LN}(X), \textsc{LN}(X)\big) \\
  \hat{T}_l(X) &= T_l(X) + f_{\text{MLP}}\big(\textsc{LN}(T_l(X))\big) \\
  T(X)    &= (\hat{T}_{L-1} \circ \hat{T}_{L-2} \circ \ldots \circ \hat{T}_0)(X).
\end{align}
In Eq.~\eqref{eq:sincos}, we concatenate, \ie$\parallel$, a sinusoidal positional encoding~\cite{vaswani2017attention} $\mathbf{z}_n^{\mathbf{X}_R} \in \mathbb{R}^{256}$ to each patch embedding to encode spatial information. Next, we cluster the patches in $\hat{\mathbf{B}}^{\mathbf{X}_R}$ for each resolution $R$ using $k$-means clustering~\cite{kmeans}:
\begin{align}
C^R_k = & \underset{C}{\operatorname{argmin}} \sum_{i=1}^{N_R} \min_{\mu_j \in C} \| \hat{\mathbf{B}}^{\mathbf{X}_R}_i - \mu_j \|^2 
\end{align}
where $C^R_k$ is the set of $k$ clusters for resolution $R$ and $\mu_j$ are the cluster centroids. Running the clustering procedure with $k \in \{2,\ldots,8\}$ on two different image resolutions results in 14 unique cluster assignments.

\boldparagraph{Cross-clustering Consistency.} Each run of $k$-means clustering labels its clusters independently from  others, yielding a correspondence mismatch between clusters across runs. To resolve this, we relabel the cluster indices to a common reference frame with the following steps:\\
\boldparagraph{1.} Select $C$ with the most clusters after $k$-means as a reference set \( S \). As some clusters end up empty during the $k$-means iterations, this is not always the set with highest initial $k$. The reference set determines the indices used for all other sets of clusters \( \mathcal{C} \), each with its number of clusters denoted by \( |C_i| \):
\vspace{-5pt}
\begin{align}
  S = \underset{C_i \in \mathcal{C}}{\mathrm{argmax}} \, |C_i|
\end{align}
\boldparagraph{2.} Sets of clusters are aligned to the reference set using Hungarian matching~\cite{hungarianmatching}. We calculate pairwise Intersection over Union (IoU) between the clusters from \( S \) and \( C \). Then, each cluster from \( C \) is assigned a new index, matching the cluster with the highest IoU from \( S \):
  \begin{align}
    &\text{For each cluster } c_j \in C, \quad j \in \{1, \ldots, |C|\}: \nonumber\\
    &\text{Assign index } i \text{ to } c_j \text{ where } i = \!\underset{i \in \{1, \ldots, |S|\}}{\mathrm{argmax}} \, \text{IoU}(c_j, s_i)
  \end{align}
\boldparagraph{3.} With the labeled sets of clusters, a probability distribution over the clusters is assigned to each image patch. For a given patch $p$, let $\mathbf{L}(p) = \{\mathbf{L}_{1}(p), \mathbf{L}_{2}(p), \ldots,\\ \mathbf{L}_{m}(p)\}$ be the set of labels assigned to $p$ by the $m$ different sets of clusters. The probability $P(n|p)$ of $p$ being assigned to a particular cluster $n$ is defined as the relative frequency of $n$ among labels $\mathbf{L}(p)$:
\vspace{-0.5em}
\begin{align}
    P(n|p) = \frac{\sum_{i=1}^{m} \mathbbm{1}}{m}, \text{\small with } \mathbbm{1} =
    \begin{cases}
            1, & \text{\small if } \mathbf{L}_{i}(p) = n \\
            0, & \text{\small else}
    \end{cases}.
\end{align}
The predictions of a single $k$-means predictor tend to contain high levels of noise, likely due to the high dimensionality of the embeddings. Our method can be seen as an ensemble, reducing the variance present in each individual predictor. The areas that are consistently clustered together by various predictors are likely to be semantically connected. In addition, our method enables for a flexible number of output clusters. Some of the initial clusters can disappear if they are not well-supported by multiple predictors. This property is highly desired due to the variety of input images and the number of objects in them.

\boldparagraph{Cluster Denoising.} 
To further improve the locality and semantic meaningfulness of clustered feature representations, we apply a Conditional Random Field (CRF)~\cite{crf} and majority filter. CRF is a discriminative statistical method that is used to denoise predictions based on local interactions between them. In our case, the predictions are a 2D grid of cluster assignment probabilities of the image patches. Our implementation is specifically tailored for refining 2D segmentation map, using a mean field approximation with a convolutional approach to iteratively adjust the probability distributions of each image patch's cluster indices. In the pairwise potentials, we use a Gaussian filter to ensure spatial smoothness and consistency in the segmentation. The application of the CRF yields embeddings which are less noisy and more cohesive than the original aligned $k$-means result. To address remaining noise in the embeddings, a neighborhood majority filter is applied as a final step. For each image patch, we consider the set of patches $\mathcal{N}(i, j)$ in its square neighborhood: $\mathcal{N}(i, j) = \{(i + \delta_i, j + \delta_j) \, | \, \delta_i, \delta_j \in \{-1, 0, 1\}\}$.
The mode value from the cluster indices in that neighborhood is calculated and assigned as the new index of the central patch:
\vspace{-3pt}
\begin{align}
  \text{mode}\big(\mathcal{N}(i, j)\big) 
  = \underset{k \in K}{\mathrm{argmax}} \sum_{m\in \mathcal{N}(i, j)} \mathbbm{1}_{\text{index}(m) = k}
\end{align}
This step is applied recursively until convergence or 8 times at most. In the supplementary material, we visualize the effect of each step on the embeddings.\\
\boldparagraph{Captioning.} The next step involves turning clustered, denoised and enhanced embeddings into text. The BLIP text decoder is a transformer architecture capable of processing unordered sets of embeddings of arbitrary size. We leverage this feature and feed flattened subsets of patch embeddings, each corresponding to a cluster, to the text decoder. Spatial information is preserved due to the presence of positional embeddings added in the clustering step. With this technique, our method essentially infers semantic categories captured by clusters and represented by BLIP embeddings. To the best of our knowledge, we are the first to use the text decoder in this manner, enabling local captioning without specifically training for it. The caption generation is stochastic, with different object namings appearing in the captions depending on initialization. To obtain a rich, unbiased and diverse set of object names, we regenerate captions with each embedding with multiple inference \textit{cycles}.\\
\boldparagraph{Caption Filtering.} The captions generated in the previous step are sentences in natural language. For our task, we are only interested in the class names present in each sentence. To obtain these, we filter the sentence down to relevant class names by extracting all nouns using part-of-speech labels for each word in the caption. Nouns are kept and converted into their singular form through lemmatization. We collect all nouns generated by different clusters and cycles into one target vocabulary and remove any duplicates, as well as nouns which do not appear in the WordNet dictionary.
\subsection{Segmentation through Self-Guidance} \label{selfguidancesect}
The output of the clustered BLIP embeddings is a 32x32 grid with cluster index assignments (see Sect.~\ref{captioning}), with each element corresponding to an image patch from the original image. This enables the extraction of segmentation masks - defined as a union of areas covered by image patches with the same index - essentially for free. For instance, Fig.\ref{fig:pipeline} and~\ref{fig:abl1} show the clustered output as a 2D mask that partially captures relevant objects. However, the low resolution of the segmentations are unsatisfactory, and upsampling leads to oversegmented objects or unsharp boundaries. To perform effective and accurate auto-vocabulary semantic segmentation, we leverage \blip's strength in generating an elaborate set of relevant class names from clusters instead - and use this as textual guidance for a pre-trained OVS model capable of producing high-resolution outputs. \blip is model-agnostic, allowing our approach to integrate with any OVS model that accepts an image and a set of class labels. In this work, we focus on X-Decoder~\cite{xdecoder}, a popular and well-performing OVS model.

\section{Evaluation of Auto-Classes} \label{llmeval}
As discussed in Sect.~\ref{relatedwork}, previous works in OVS have mainly focused on a setting where the target class names are provided by the user. Hence, evaluation is possible by having access to the ground truth of those target class names during evaluation. However, in scenarios where the target class names are discovered, as in our framework, rather than pre-specified there may be a lack of direct alignment between the semantics of these categories and the classes used in the annotations. In zero-guidance segmentation~\cite{zeroguidance}, the authors have proposed to align class names based on the cosine similarity in the latent space of Sentence-BERT~\cite{sentencebert} or CLIP~\cite{clip}. In our initial tests, however, this approach misaligned obvious class pairs (e.g., \textit{taxi} mapped to \textit{road} instead of \textit{car}), thereby reducing segmentation accuracy despite promising qualitative results. Relations between two class names can be complex and ambigious~\cite{YADAV202185}, such as due to synonymity, hyponymy, or hypernymy, which exceed the capabilities of a cosine-similarity criterium in the latent space. To address this, we propose an \textbf{L}LM-based \textbf{A}uto-\textbf{V}ocabulary \textbf{E}valuator (LAVE), leveraging the Llama-2-7B language model~\cite{llama}, to map predicted auto-vocabulary categories to target dataset classes. LAVE gathers all predicted auto-vocabulary classes and maps each category to the most relevant or similar class in the known vocabulary. It then updates all pixel values in the predicted segmentation masks according to the mapping, after which the mean Intersection over Union (mIoU) is calculated using the updated mask. Though not an integral part of our method, LAVE greatly reduces the mapping effort which is often infeasible to do manually, requiring $C \times A$ comparisons between $C$ known classes and $A$ auto-classes. Pseudocode and prompts for LAVE are included in the supplementary material.

%% file: Sections/4_experiments.tex
\begin{figure*}[t!]
  \centering
  \includegraphics[width=0.95\linewidth]{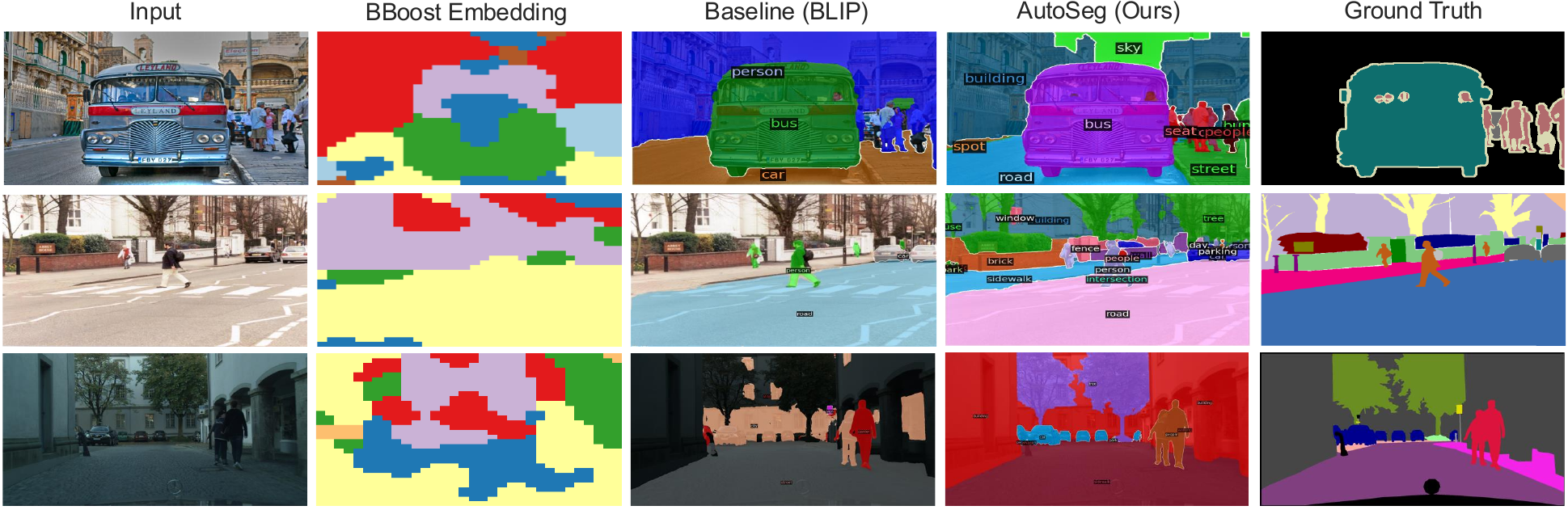}\\
  \caption{\textbf{Segmentation with VLMs.} Example outputs on PASCAL VOC/Context (top), ADE (middle) and Cityscapes (bottom) by (left to right) directly using BBoost embeddings as masks, feeding plain BLIP embeddings to X-Decoder or \ours. Notably, our method segments images in the most comprehensive and semantically accurate manner.}
  \label{fig:abl1}
  \vspace{-10pt}
\end{figure*}
\input{Tables/vlm}
\section{Experiments}
\subsection{Experimental Setup}
\boldparagraph{Datasets.}
\ours is evaluated on four popular semantic segmentation validation datasets: PASCAL VOC~\cite{VOC} and Context (PC)~\cite{context}, ADE20K (ADE)~\cite{ADE20k} and Cityscapes (CS)~\cite{cityscapes} with 20, 459, 847 and 20 classes respectively, covering a wide range of difficulty and class diversity. For instance, ADE is challenging with its many and infrequently appearing classes, while CS contains many instances per image (see Tab.~\ref{tab:blip_tab} for detailed statistics).\\
\boldparagraph{Evaluation.}
For quantitative comparison with previous works, we use mIoU as the main metric. As mentioned in Sec.~\ref{llmeval}, auto-class predictions in the segmentation mask are mapped using LAVE before the mIoU is computed. Furthermore, to study the class-agnostic segmentation performance of different VLMs, we report class-agnostic mean Intersection over Union (cmIoU). The computation of cmIoU involves the following steps. First, pairwise IoU values, denoted as $\text{IoU}(g_i, p_j)$ are calculated between all ground-truth segments $g_i$ and predicted segments $p_j$ from a given image. Each $g_i$ segment is matched with the predicted segment $p_{\text{max}_i}$ that yields the highest IoU value: $p_{\text{max}_i} = \arg \max_j \text{IoU}(g_i, p_j)$. The cmIoU is then determined by calculating the arithmetic mean of these highest IoU values across the dataset, which can be formulated as $\text{cmIoU} = \frac{1}{N} \sum_{i=1}^{N} \text{IoU}_{\text{max}_i}$ where $N$ is the total number of ground-truth segments in the dataset.\\
\boldparagraph{Implementation details.}
For our experiments, we use the BLIP model built with ViT-Large backbone and finetuned for image captioning on the COCO dataset in combination with the Focal-L variant of X-Decoder. For both models, we use publicly available pretrained weights and do not perform any additional finetuning. Parameter tuning is performed for the clustering and captioning modules. Parameters of the clustering include image scales encoded by BLIP ($384\times384$ and $512\times512$), the $k$ values of $k$-means clustering (2 to 8), the parameters of Gaussian smoothing in the CRF (smoothness weight of 6 and smoothness $\theta$ of 0.8), the number of iterations of majority filtering (8) and the feature dimension size of the positional embeddings (256). For text generation, we use nucleus sampling with a minimum length of 4 tokens and maximum of 25, top P value of 1 and repetition penalty of 100 to ensure as many unique nouns as possible. These parameters were determined using Bayesian optimization on VOC with the goal of maximizing the mIoU. For CRF denoising and part-of-speech tagging, we use the crfseg~\cite{crf} and spaCy~\cite{spacy} libraries respectively.

\subsection{Ablations}
To assess our framework we investigate various segmentation methods with captions, the impact of captioning cycles, generated/fixed vocabulary similarity, alternative mappers and the effects of cluster denoising. The latter three ablation studies are detailed in the supplementary material.

\boldparagraph{Segmentation with Vision Language Models.} Upsampled \blip embeddings could potentially act as segmentation masks on their own given the clustering of semantic areas (see Sec.~\ref{selfguidancesect}). In this ablation study, we compare this baseline to \ours which utilizes \blip's low-resolution semantics for high-resolution outputs. We assess segmentation quality using class-agnostic (c-mIoU), with results presented in Tab.~\ref{tab:vlm} and Fig.~\ref{fig:abl1}. Across all datasets, \ours performs best, with the baseline X-Decoder + BLIP showing issues of false positives and negatives that explain the performance gap. For example, it incorrectly segments areas around objects in VOC/PC (\eg \textit{person} for \textit{building}) and ignores significant regions in ADE and CS. In contrast, \ours effectively captures smaller objects and details. As expected, low-resolution \blip masks are inaccurate but can still segment general areas in VOC/PC and ADE, though they struggle in CS due to higher object density.

\input{Tables/captioning_cycles}

\boldparagraph{Captioning Cycles.} A key aspect of our model is its capability to repeatedly enrich the vocabulary with \blip, enabling it to generate a more exhaustive collection of classes which accurately describe the scene. This approach proves especially beneficial for instances where objects are initially overlooked or described with less precise semantic terms. To assess its impact, we explore the effects of various captioning iterations using the mIoU metric. The findings, presented in Tab.~\ref{tab:blip_tab}, reveal that the optimal number of captioning cycles varies significantly with the dataset in question. Specifically, we observe that for CS, characterized by its high number of instances and unique classes per image (denoted as $\overline{I}$ and $\overline{C}$ in Tab.~\ref{tab:blip_tab}, respectively), a greater number of cycles yields the most benefit. For datasets with a lower average number of instances and unique classes per image, like VOC or ADE, fewer captioning cycles are required for \blip to identify all relevant class names. As anticipated, our model requires a greater number of captioning cycles for the CS dataset given its additional 439 object categories. Remarkably, \ours successfully handles ADE, the dataset with the largest number of unique classes in our experiments, using merely a single captioning cycle. This outcome underscores our model's ability in efficiently addressing complex datasets which resemble an open-ended setting. Finally, we note that through its captioning framework, \ours perceives significantly more distinct classes compared to the hand-crafted fixed vocabularies.

\subsection{Quantitative Analysis}
\boldparagraph{Auto-Vocabulary Segmentation Setting.} Our quantitative analysis begins by comparing \ours with other auto-vocabulary methods, such as Zero-Guidance Segmentation (ZeroSeg~\cite{zeroguidance}), as well as LLM-aided segmentation methods like LISA~\cite{lisa} with alternative captioning methods like LLaVA~\cite{llava}. We also evaluate a configuration combining BLIP and X-Decoder, where instance crops predicted by SAM~\cite{SAM} are used for captioning to simulate locality. We use the same caption filtering as in our method. The results, shown in Tab.~\ref{tab:rebut}, indicate that our method outperforms ZeroSeg, the only other true auto-vocabulary approach. In contrast, LISA requires multiple inferences with individually prompted categories to achieve scene segmentation—a notably more complex and less efficient approach that also yields poor performance. Captioning based on smaller object crops is similarly ineffective compared to using enhanced vision-language features that maintain spatial locality. Moreover, while LISA provides LLM-guided captions, these lack the contextual specificity needed to establish an effective vocabulary for X-Decoder. Although \ours also utilizes this segmentation backbone, it achieves a substantially higher performance. This comparison not only highlights the unique strengths of \ours but also reveals the practical limitations of current LLM-aided approaches in segmenting scenes with multiple object categories.
\input{Tables/auto_gen}
\begin{figure*}[ht!]
  \centering
  \includegraphics[width=0.89\linewidth,height=8cm]{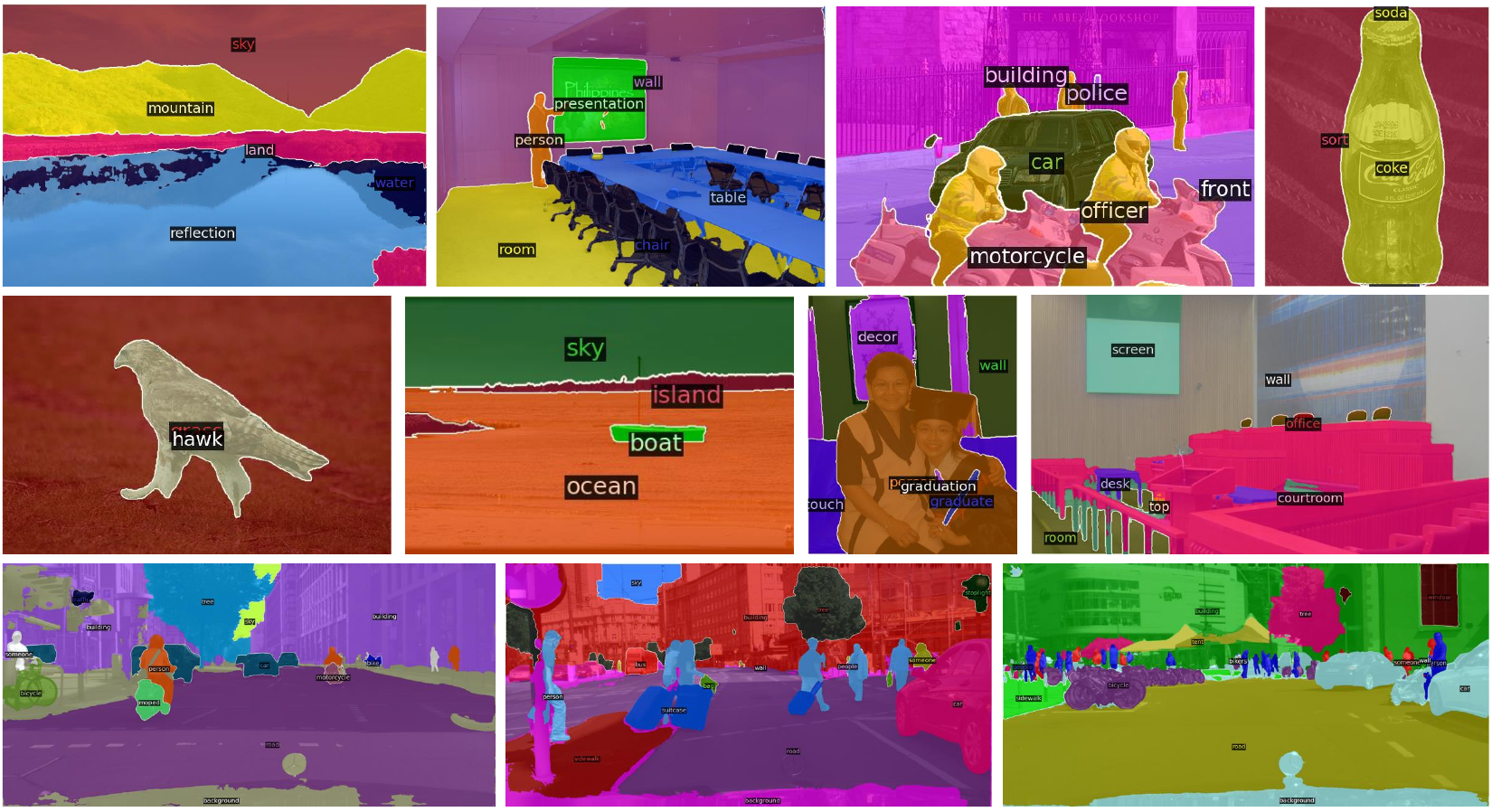}\\
  \caption{\textbf{Qualitative Results.} \ours shows remarkable capability to identify out-of-vocabulary categories, such as \textit{hawk} or \textit{coke}, and segment them accurately across different datasets. Images are from the VOC/PC, ADE and CS datasets.}
  \label{fig:qualitative}
  \vspace{-6pt}
\end{figure*}
\input{Tables/quantitative}\\
\boldparagraph{Open-Vocabulary Segmentation Setting.} We compare our method against existing OVS methods that require class names to be given. Tab.~\ref{tab:quantitative} shows the results. In addition to the results mapped with LAVE, we provide results with one manual mapper per dataset (note that we provide a manual mapper for this table only, given the significant manual effort to construct it). Without any specification of class names through user input, \ours matches 91\%, 67\%, 42\% and 55\% of the best OVS method performance on VOC, PC, ADE and CS respectively. It should be noted that this metric reflects the performance on the \textit{known}, annotated classes, while additional open-ended classes are potentially mapped. Despite not being explicitly instructed with the known vocabulary contrary to OVS methods, \ours is remarkably able to surpass six of them on VOC and PC, such as OpenSeg~\cite{openseg}, ODISE~\cite{odise} or OVSeg~\cite{ovseg}. While still leaving room for improvement, our model compares competitively with OVS methods on ADE and CS, two very challenging datasets either high in number of unique classes or instances. Compared to ZeroSeg, \ours achieves superior performance on VOC (88.2 over 20.1 mIoU) and PC (12.8 over 11.4 mIoU). Furthermore, we set the first benchmark on ADE and CS under the unknown vocabulary task setting. This outcome underscores the efficacy of our method in dealing with complex scenes that are open-ended in nature, such as ADE with its large number of rare classes. Finally, results obtained with LAVE mappings show little difference with manually constructed ones. This indicates that at marginal cost, LAVE can act as a feasible bridge between a known and unknown vocabulary. Our method deals well with high numbers of instances and can successfully identify various non-salient objects. While \blip struggles with producing high-quality masks by itself (as seen in Tab.~\ref{tab:blip_tab}), it is highly effective in embedding target classes accurately which can be segmented with precision afterwards. Overall, our results demonstrate that the integration of \blip with OVS harnesses the strengths of both, leading to enhanced open-ended recognition capabilities.

\subsection{Qualitative Analysis}
Fig.~\ref{fig:qualitative} displays qualitative results on the four datasets, where \ours demonstrates its capability to accurately detect and segment relevant object categories present in the images but missing from the ground truth vocabulary. Remarkably, it predicts segmentation masks for classes such as \textit{moped}, \textit{presentation}, \textit{coke}, \textit{courtroom} or \textit{hawk}. Moreover, in certain instances, \ours successfully captures additional contextual details, such as \textit{graduation} or \textit{reflection}, illustrating the model's capabilities at semantic segmentation in a genuinely open-ended manner. Additional results, including failure cases, are included in the supplementary material.

%% file: Tables/vlm.tex
\begin{table}[tb]
\caption{\textbf{Ablations on Segmentation with VLMs (c-mIoU).} Leveraging \blip semantic embeddings for segmentation with \ours outperforms baselines on all four datasets.}
\centering
\scriptsize
\setlength{\tabcolsep}{6pt}
\renewcommand{\arraystretch}{1.2}
\newcommand{\csp}{\hskip 2em}
\begin{tabular}{l ccccc}
\toprule
\textbf{Method} & \textbf{VOC}~\cite{VOC} & \textbf{PC}~\cite{context} & \textbf{ADE}~\cite{ADE20k} & \textbf{CS}~\cite{cityscapes} \\
\midrule
BBoost Embeddings                            & \rd 16.3 & \rd 16.3 & \rd 11.3 & \rd 0.85\\
X-Decoder~\cite{xdecoder} + BLIP~\cite{blip} & \nd 38.0 & \nd 35.3 & \nd 26.7 & \nd 29.2 \\
\ours (Ours) & \fs 71.8 & \fs 47.7 & \fs 29.2 & \fs 35.8 \\
\bottomrule
\end{tabular}
\label{tab:vlm}
\vspace{-10pt}
\end{table}

%% file: Tables/captioning_cycles.tex
\begin{table}[t]
  \caption{\textbf{Ablations on Captioning Cycles (mIoU) and Generated Classes.} The optimal number of captioning cycles depends on the dataset, influenced by the image content and annotation method. Our method identifies a substantially larger variety of unique classes (\textbf{Gen.}) than those annotated by humans (\textbf{Ann.}), highlighting its capacity to capture finer context. $\overline{I}$ and $\overline{C}$ denote the average number of instances and classes per image.}
  \label{tab:blip_tab}
  \centering
  \scriptsize
  \setlength{\tabcolsep}{2pt}
  \renewcommand{\arraystretch}{1.2}
  \newcommand{\csp}{\hskip 2em}
  \begin{tabular}{l rrrrrr|rr rr}
    \toprule
    \multicolumn{1}{c}{} & \multicolumn{6}{c}{\textbf{Number of Captioning Cycles}} & \multicolumn{4}{c}{\textbf{Data Properties}} \\[3px]
    \textbf{Dataset} & \textbf{1} & \textbf{5} & \textbf{10} & \textbf{15} & \textbf{20} & \textbf{25} & \textbf{Gen.} & \textbf{Ann.} & $\overline{I}$ & $\overline{C}$ \\
    \midrule
    PASCAL VOC~\cite{VOC} & \rd 85.3 & 84.4 & \fs 87.1 & \nd 85.7 & 83.3 & 83.3 & 938 & 20 & 3.5 & 3.5\\
    PASCAL Context~\cite{context} & 10.1 & 9.7 & \nd 11.2 & \fs 11.7 & 9.6 & \rd 10.8 & 721 & 459 & 18.9 & 6.2\\
    ADE20K~\cite{ADE20k} & \fs 6.0 & \nd 5.8 & 5.1 & 5.2 & 5.1 & \rd 5.6 & 1578  & 847 & 19.5 & 10.5\\
    Cityscapes~\cite{cityscapes} & 28.2 & 28.1 & \nd 28.4 & \rd 28.3 & 27.9 & \fs 30.0 & 395 & 20 & 34.3 & 17\\
    \bottomrule
    \end{tabular}
\vspace{-10pt}
\end{table}

%% file: Tables/auto_gen.tex
\begin{table}[t]
   \caption{\textbf{Quantitative Auto-Vocabulary Results (mIoU).} The wide availability of captioning and OVS methods allows various combinations to design an auto-vocabulary method, but they are not necessarily performing well. \ours performs superior over the only Auto-Vocabulary method ZeroSeg, as well as over AVS-adapted methods with alternative captioners or segmentors.}
   \centering
\scriptsize
\setlength{\tabcolsep}{6pt}
\renewcommand{\arraystretch}{1.2}
\newcommand{\csp}{\hskip 2em}
  \begin{tabular}{lrrr}
    \toprule
    \textbf{Auto-Vocabulary Segmentation Method} & \textbf{VOC}~\cite{VOC} & \textbf{PC}~\cite{context} &\textbf{CS}~\cite{cityscapes} \\
    \midrule
    LLaVA + LISA~\cite{llava, lisa}                   &  7.7 & 0.2 & 1.5 \\
    ZeroSeg~\cite{zeroguidance}                       & 20.1 & \nd 11.4 & - \\
    SAM + BLIP + X-Decoder~\cite{SAM, blip, xdecoder} & \rd 41.1 & \rd 11.3 & \nd 27.4 \\
    LLaVA + X-Decoder~\cite{llava, xdecoder}          & \nd 56.7 & \nd 11.4 & \rd 23.4 \\
    \ours (Ours)                                      & \fs 87.1 & \fs 11.7 & \fs 30.0 \\
    \bottomrule
  \end{tabular}
  \label{tab:rebut}
  \vspace{-10pt}
\end{table}

%% file: Tables/quantitative.tex
\begin{table}[tb]
  \caption{\textbf{Open/Auto-Vocabulary State of the Art Comparison (mIoU).} \ours surpasses ZeroSeg~\cite{zeroguidance} in human-free segmentation and remains competitive with some well-known OVS methods that rely on human input, while outperforming others (3 out of 8 methods on VOC and 3 out of 7 methods on PC). Results with dashes indicate unpublished or unavailable data.}
  \centering
  \scriptsize
  \setlength{\tabcolsep}{2.1pt}
  \renewcommand{\arraystretch}{1.1}
\begin{tabular}{@{}l@{\hspace{-7pt}}crrrr}
    \toprule
        & \textbf{Unknown} & \textbf{VOC}~\cite{VOC} & \textbf{PC}~\cite{context} & \textbf{ADE}~\cite{ADE20k} & \textbf{CS}~\cite{cityscapes} \\
    \textbf{Method} & \textbf{Vocabulary} & (20) & (459) & (847) & (20)\\
    \midrule
    \rowcolor{gray!20} \multicolumn{6}{l}{OVS Segmentation methods with prompted (\emph{known} = ground truth) vocabulary}\\
    LSeg~\cite{lseg}               & \noo & 47.4 &   -  &   -  & - \\
    OpenSeg~\cite{openseg}         & \noo & 72.2 & 9.0 &  8.8 & - \\
    OVSeg~\cite{ovseg}             & \noo & 94.5 & 11.0 &  9.0 & - \\
    ODISE~\cite{odise}             & \noo & 82.7 & 13.8\rd & 11.0 & - \\
    SAN~\cite{san}                 & \noo & 94.6 & 12.6 & 12.4\rd & - \\
    CAT-Seg~\cite{catseg}          & \noo & \textbf{97.2}\fs & \textbf{19.0}\fs & 13.3\nd & - \\
    X-Decoder~\cite{xdecoder}      & \noo & 96.2\nd & 16.1\nd &  6.4 & 50.8\nd \\
    FC-CLIP~\cite{fcclip}          & \noo & 95.4\rd & 12.8 & \textbf{14.8}\fs & \textbf{56.2}\fs \\
    \midrule
    \rowcolor{gray!20} \multicolumn{6}{l}{AVS Segmentation methods with auto-generated (\emph{unknown}) vocabulary}\\
    ZeroSeg~\cite{zeroguidance}   & \yes & 20.1\rd & 11.4\rd & - & - \\
    \ours + LAVE mapper (Ours) & \yes & \nd 87.1 & 11.7\nd & 6.0\nd & 30.0\nd \\
    \ours + Manual mapper (Ours) & \yes & \bf 88.2\fs & \bf 12.8\fs & \bf 6.2\fs & \textbf{31.1} \fs \\
    \bottomrule
  \end{tabular}
  \label{tab:quantitative}
\vspace{-10pt}
\end{table}

%% file: Sections/5_conclusion.tex
\section{Conclusion}
This paper introduced \ours, a novel method which leverages a vision-language model to automatically generate relevant target classes and segment them. Additionally, we proposed \textsc{LAVE}, a new evaluation framework which maps open-ended class names to ground-truth labels. \ours shows open-ended recognition capabilities, achieving state-of-the-art performance in the zero label setting, while being competitive with open-vocabulary segmentation models which require provided ground-truth labels.

%% file: Sections/6_acknowledgements.tex
\boldparagraph{Acknowledgements.}
This work was supported by TomTom, the University of Amsterdam and the allowance of Top consortia for Knowledge and Innovation (TKIs) from the Netherlands Ministry of Economic Affairs and Climate Policy.

%% file: Sections/x_supplementary.tex
\section{Ablation on Denoising and Caption Filtering}\label{appendix:A}
This section aims to provide more insight into the effects of denoising and caption filtering on the final performance. Additionally, we analyze the effects of each enhancement step by visualizing the influence on the vision-language embeddings.

\subsection{Effects on Performance}\label{appendix:B}
Our enhancement of the original BLIP embeddings involves iterative semantic clustering followed by denoising using a Conditional Random Field (CRF) and majority filter. As shown in Table~\ref{tab:bboost}, performing denoising after clustering significantly improves performance, increasing mIoU from 70.4 to 87.1 on PASCAL VOC~\cite{VOC}, confirming the effectiveness of this denoising step. Additionally, we analyze the impact of caption filtering by evaluating mIoU performance with this step disabled. The results in Table~\ref{tab:bboost} demonstrate that caption filtering is essential, regardless of the enhancements applied to the original BLIP embeddings. This emphasizes the importance of selecting relevant words from the decoded text.

\subsection{Visualization of Embeddings in BBoost} \label{appendix:C}
\begin{figure*}[t!]
  \centering
  \includegraphics[width=\linewidth]{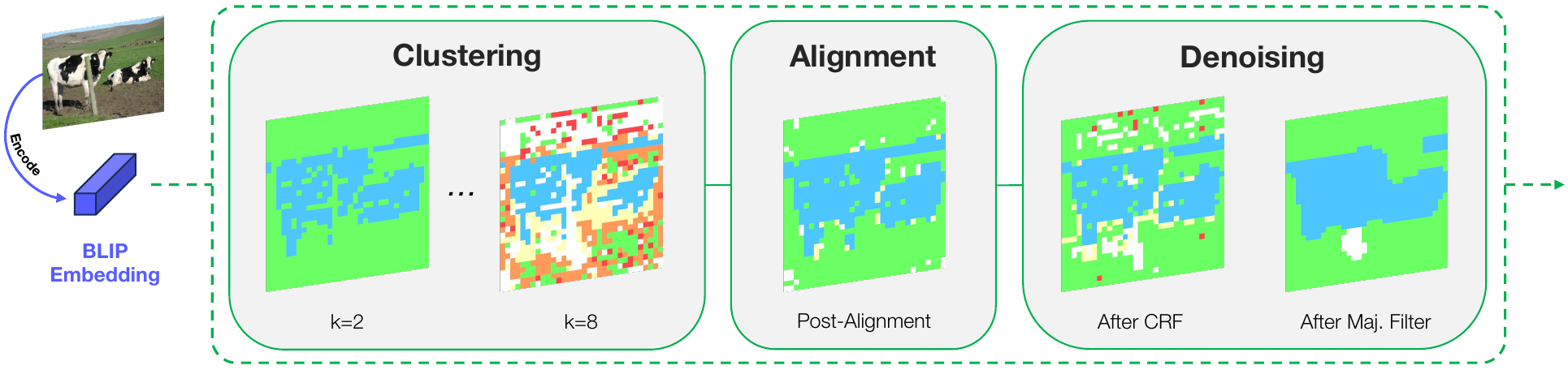}\\
  \caption{\textbf{Visualization of Intermediate BBoost Outputs.} Intermediate BBoost embeddings, visualized as 32x32 grids, show how vision-language features are gradually enhanced into semantically regions that are more coherent, which proved more effective for text generation that preserves locality. After the last step, the embeddings are decoded into text by the text decoder.}
  \label{fig:bccembeddings}
\end{figure*}

In Fig.~\ref{fig:bccembeddings}, we visualize intermediate 32x32 outputs of BBoost to better understand what each individual step of clustering, alignment and denoising achieves. Firstly, Fig.~\ref{fig:bccembeddings} reveals that clustering BLIP embeddings results grouping pixels into semantic areas. With $k=2$, the image is divided an area which includes the cows (in blue) and another for everything else (green), while for $k=8$, various other objects - such as the pole in purple - get included. At this stage, the embeddings appear noisy and inconsistent across all $k$, which is why the combination of all clusters is so important. The alignment output shows that certain objects from a higher $k$ embeddings are now included in those of lower $k$ embeddings, \ie the aligned embedding captures a more extensive list of objects present in the image. Finally, to further reduce the noise in the embedding, we apply CRF and a majority filter. We see that the final embedding consists of three unique semantic areas, which are eventually fed to the text decoder for text generation.

\section{Generated/Fixed Vocabulary Similarity}\label{appendix:D}
To investigate how similar vocabularies generated with \ours are to the fixed ones, we measure how many of the fixed classes are directly included (with identical wording)  in the generated vocabulary. Additionally, we measure the average CLIP cosine similarity between the CLIP embedding of each fixed class to the CLIP embedding of the most similar generated class. Table~\ref{tab:vocabularyaccuracy} shows the results. We find that there is a higher direct overlap with smaller vocabularies, which can potentially be explained by the specificity of classes in datasets with larger fixed vocabularies. However, when reflecting on the average CLIP cosine similarity, it becomes clear that the generated and fixed vocabulary share high similarity across all datasets. Overall, we conclude that the coverage of the annotated classes of the four datasets is sufficient across the generated vocabularies. Similarly, the generated classes are related to the annotated ones to a great extent.

\renewcommand{\thetable}{A\arabic{table}}
\begin{table}
\centering
\scriptsize
\setlength{\tabcolsep}{3pt} 
\renewcommand{\arraystretch}{1.2}
\newcommand{\csp}{\hskip 2em}
\caption{\textbf{Similarity between generated and fixed vocabularies.} Generated vocabularies include classes that sufficiently overlap with annotated classes.}
\begin{tabular}{l cccc}
\toprule
\textbf{Dataset} & \textbf{Number of Classes} & \textbf{Generated} & \textbf{\% included} & \textbf{Avg. \text{CLIP}-sim} \\
\midrule
VOC & 20  & 938 & 70.0\% & 85.3\% \\
PC  & 459  & 669 & 37.7\% & 91.5\% \\
ADE  & 847  &  913 & 41.3\% & 92.0\% \\
Cityscapes  & 20  & 380 & 75.0\%& 85.7\% \\
\bottomrule
\end{tabular}
\label{tab:vocabularyaccuracy}
\end{table}

\input{Tables/bboost}
\input{Tables/mapper}

\section{LLM-based Auto-Vocabulary Evaluator} \label{appendix:C}
This section provides more insights into the comparison of our LLM-based Auto-Vocabulary Evaluator (LAVE) with other mappers, as well as the pseudocode and text prompts used for each respective dataset. LAVE is key to enabling the evaluation of predicted auto-vocabulary classes on public benchmarks such as PASCAL VOC~\cite{VOC}, as it allows us to compare with other auto-vocabulary or open-vocabulary methods which predict a fixed-vocabulary for evaluation.

\subsection{Mapping Comparison}
As detailed in the main paper, we initially experimented with Sentence-BERT~\cite{sentencebert} and CLIP~\cite{clip} to map auto-vocabulary classes to the fixed annotated vocabulary of the respective dataset. In this approach, we compute the text encoding for each class in the auto-vocabulary as well as the target vocabulary, and select the target class with the highest cosine similarity. However, this strategy proved ineffective, as the mapping often failed for auto-vocabulary classes with clear target matches—for example, mapping \textit{taxi} to \textit{road} instead of \textit{car}. To quantitatively evaluate this issue, we compared the mIoU performance achieved using these mappings against that of LAVE. Intuitively, poorer mappings should lead to lower pixel prediction accuracy and therefore reduced mIoU. Table~\ref{tab:mapper} demonstrates that, using the same auto-vocabulary, LAVE achieves an mIoU that is 13.2 points higher than Sentence-BERT and 9.6 points higher than CLIP. These results highlight LAVE's effectiveness in accurately identifying the correct target class, enabling a more reliable evaluation of AVS methods.

\subsection{LLM Mapper Pseudocode}
The LLM Mapper is tasked with creating a mapping dictionary which maps from auto-vocabulary classes to fixed-vocabulary classes. The resulting mapping dictionary can be used during inference to map pixel-level class predictions belonging to an automatically generated out-of-vocabulary class to fixed-vocabulary class indices. In Alg.~\ref{algorithm}, we detail the procedure to create the mapper.

\subsection{LLM Mapping Prompts}
The \textbf{generateDialogs()} function in the LLM Mapper takes a prompt template as one of its inputs. This prompt template is dependent on the dataset and is used to generate dialogs to query the LLM with. For PASCAL VOC and Cityscapes, we explicitly specify the list of categories given its short length with 20 classes each. For PASCAL Context and ADE20K, with its 459 and 849 classes respectively, we rely on the LLM's knowledge of the datasets. Explicitly naming each object class from these datasets causes the LLM to run into memory issues when queried with long dialogues. We specify the prompt templates for each respective dataset as follows:\\

\boldparagraph{PASCAL VOC and Cityscapes}\\
``To which class in the list \verb|<dataset>| is 
\verb|<name>| exclusively most similar to? If 
\verb|<name>| is not similar to any class in the 
list or if the term describes stuff instead 
of things, answer with \verb|background|. Reply 
in single quotation marks with the class name 
that is part of the list \verb|<dataset>| and do 
not link it to any other class name which 
is not part of the given list or \verb|background|." where \verb|<noun>| is the noun to map to the vocabulary and \verb|<dataset>| is the list of classes of PASCAL VOC or Cityscapes.\\

\boldparagraph{PASCAL Context and ADE20K}\\
``To which class in the \verb|<dataset>| dataset is \verb|<name>| exclusively most similar to? If \verb|<name>| is not similar to any class in \verb|<dataset>| or if the term describes stuff instead of things, answer with \verb|background|. Reply in single quotation marks with the class name that is part of \verb|<dataset>| and do not link it to any other class name which is not part of the given list or \verb|background|." where \verb|<noun>| is the noun to map to the vocabulary and \verb|<dataset>| is either `PASCAL-Context 459' or `ADE20K'.

\section{Additional Qualitative Results} \label{appendix:D}
\begin{figure}
  \centering
\begin{subfigure}{.4\textwidth}
  \includegraphics[width=\linewidth]{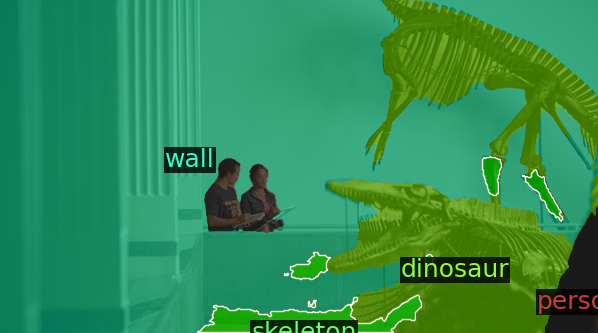}
  \caption{\textbf{Success Case.} The rare and out-of-vocabulary class name \textit{dinosaur} is generated automatically and segmented.}
  \vspace{8pt}
\end{subfigure}
\begin{subfigure}{.4\textwidth}
  \centering
  \includegraphics[width=\linewidth]{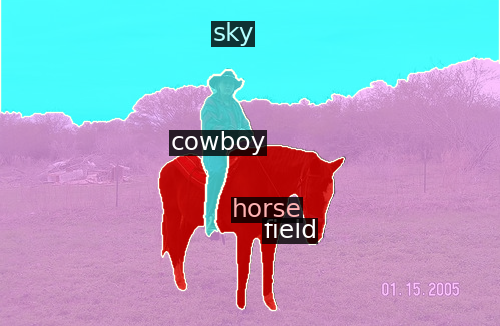}
  \caption{\textbf{Success Case.} \ours is able to accurately classify and segment classes not present in the dataset, such as the \textit{cowboy}.}
    \vspace{8pt}
\end{subfigure}
\begin{subfigure}{.4\textwidth}
  \centering
  \includegraphics[width=\linewidth]{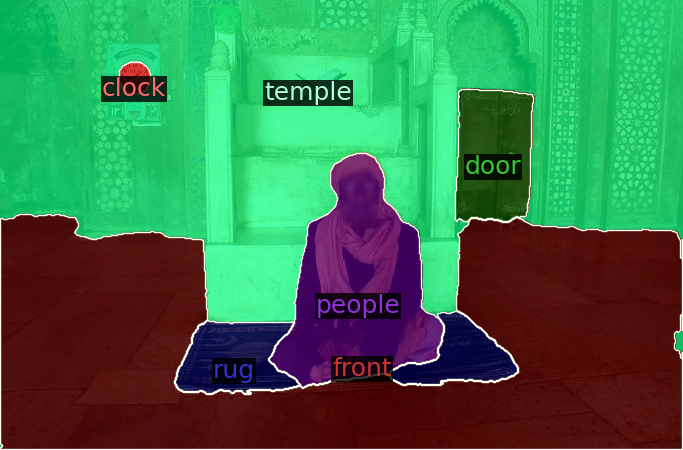}
  \caption{\textbf{Failure Case.} \ours occasionally struggles with indoor scenes from specific buildings; for example, in this case, a \textit{mosque} is mistakenly identified as a temple.}
\vspace{8pt}
\end{subfigure}
\vspace{8pt}
\begin{subfigure}{.4\textwidth}
  \centering
  \includegraphics[width=\linewidth]{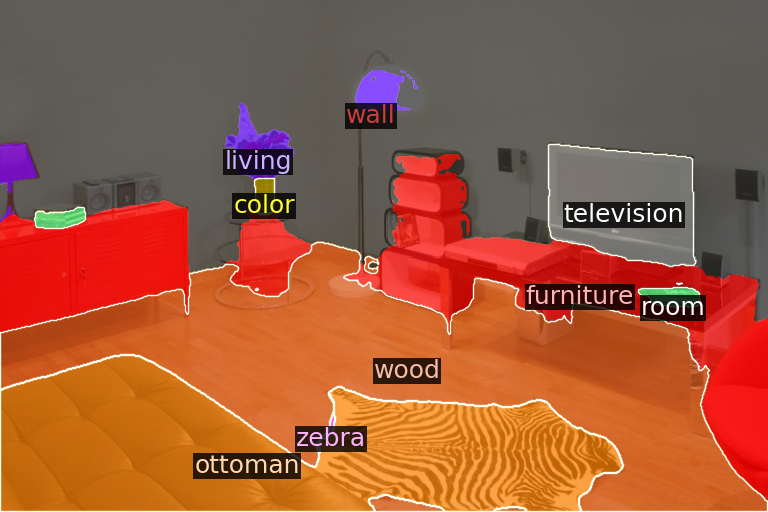}
  \caption{\textbf{Failure Case.} A \textit{rug} is mistaken for a \textit{zebra}, potentially due to \textit{rug} not being generated as a class on itself.}
\end{subfigure}
\caption{\textbf{Additional Qualitative Results}. We discuss a selection of success and failure cases on PASCAL and ADE20K above.}
\label{fig:qualitative_extra_ade}
\end{figure}

\begin{figure}[h!]
\centering
\begin{subfigure}{.4\textwidth}
  \centering
  \includegraphics[width=\linewidth]{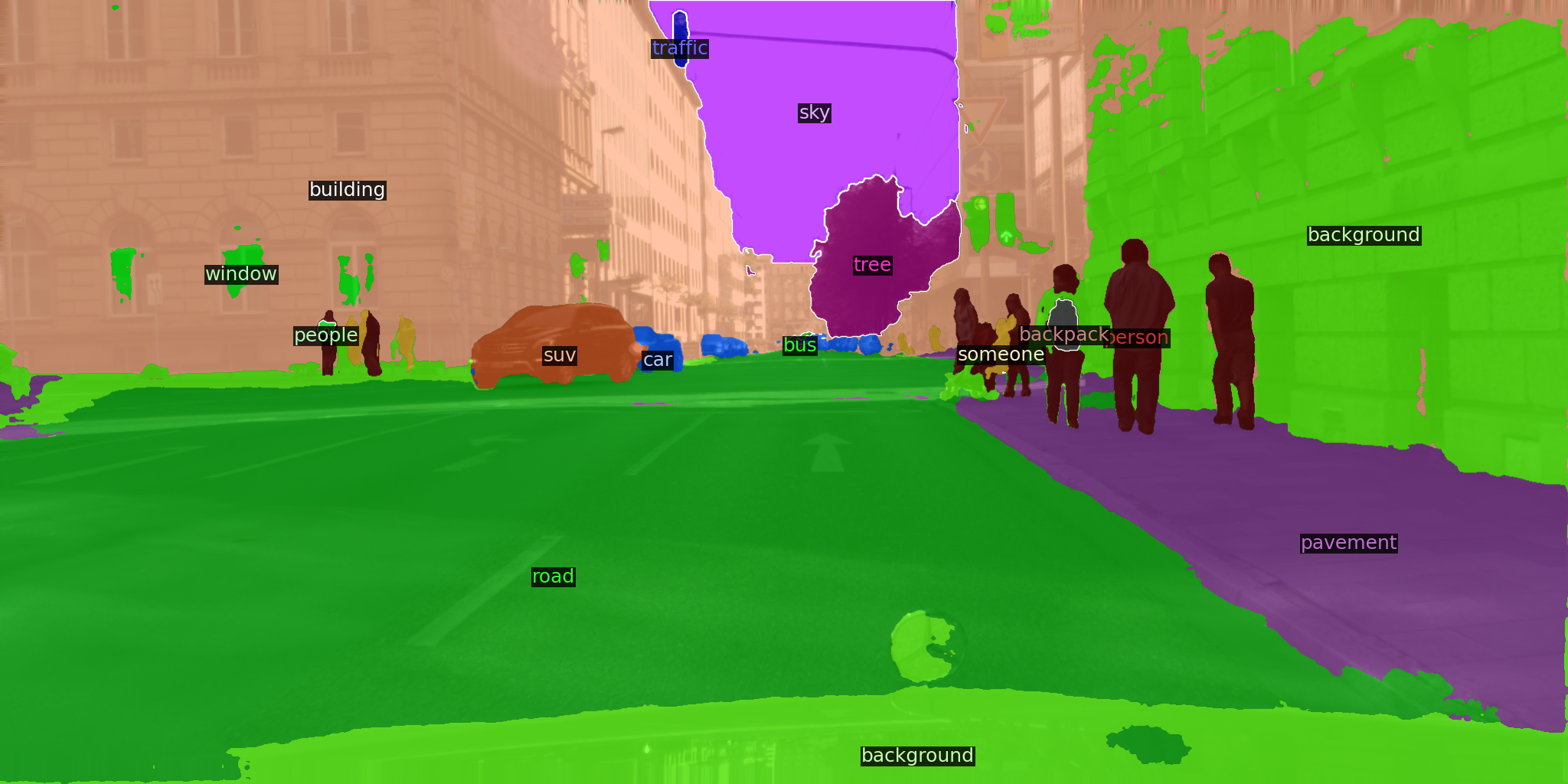}
  \caption{\textbf{Success Case.} While annotations in Cityscapes consider all car types to be equal, \ours often generates more specific descriptions, such as the \textit{SUV} in this image.}
  \vspace{8pt}
\end{subfigure}
\begin{subfigure}{.4\textwidth}
  \centering
  \includegraphics[width=\linewidth]{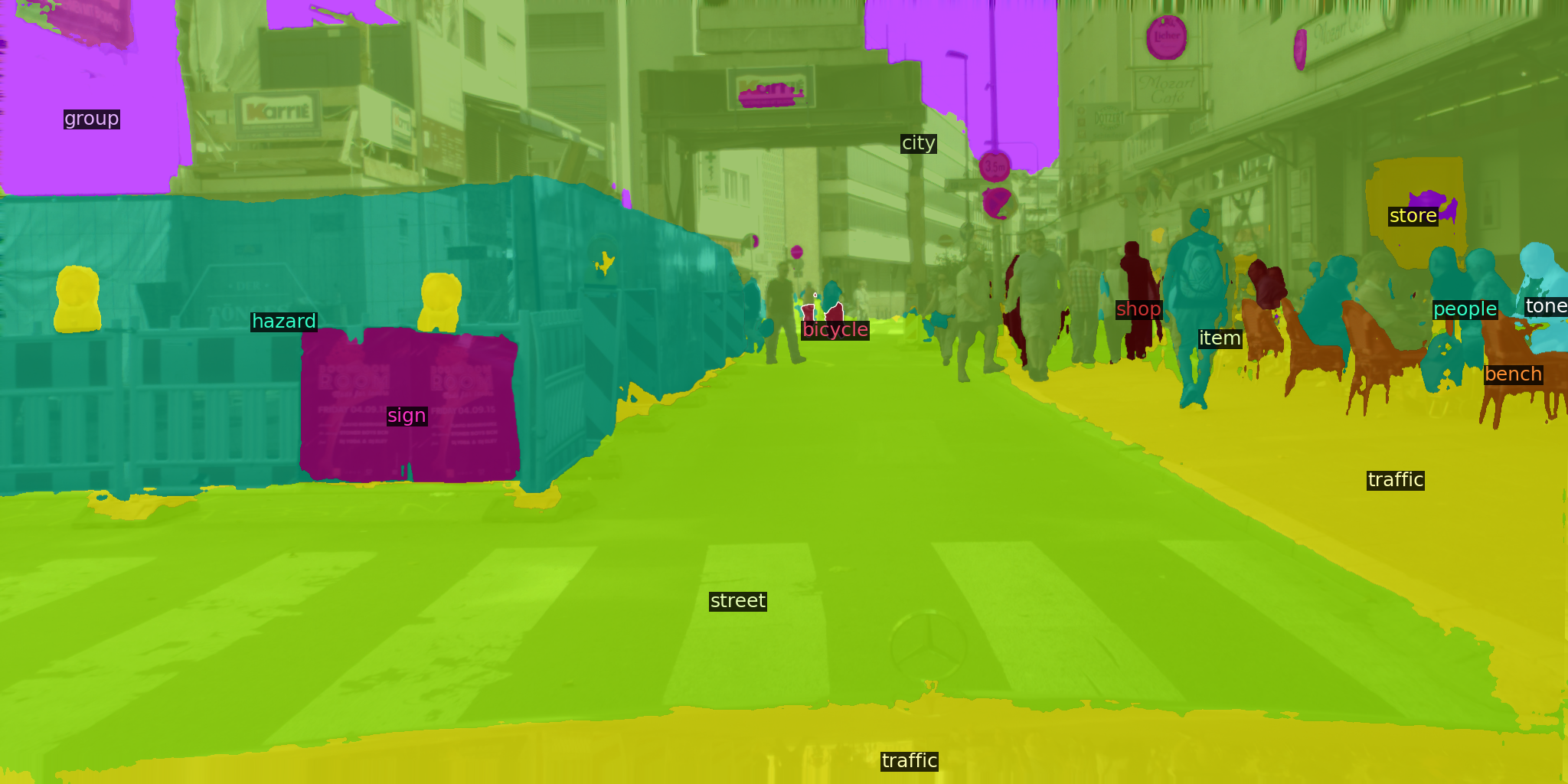}
  \caption{\textbf{Success Case.} \ours is able to provide class names which can be crucial for real-life scenarios, such as \textit{hazard} in this example. The standard 20 classes in Cityscapes do not account for such descriptions.}
\vspace{8pt}
\end{subfigure}
\begin{subfigure}{.4\textwidth}
  \centering
  \includegraphics[width=\linewidth]{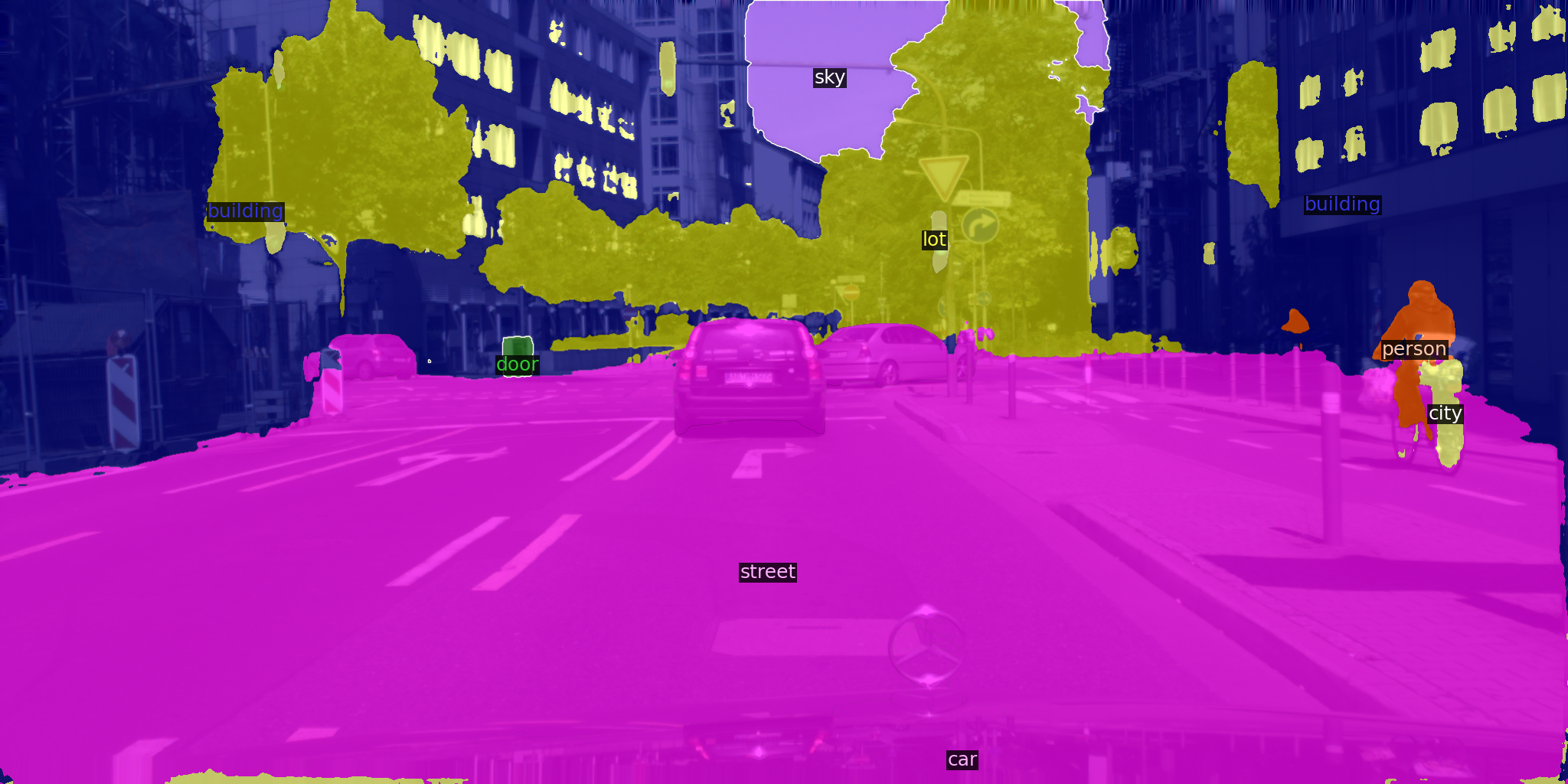}
  \caption{\textbf{Failure Case.} Occasionally \ours generates general class names, such as \textit{street}, which dominate the search space of the segmentor and cause it to attend more on that class. This may result in oversegmentation and other class names, such as \textit{car}, to be ignored. Such general classes, however, could be filtered out.}
\vspace{8pt}
\end{subfigure}
\begin{subfigure}{.4\textwidth}
  \centering
  \includegraphics[width=\linewidth]{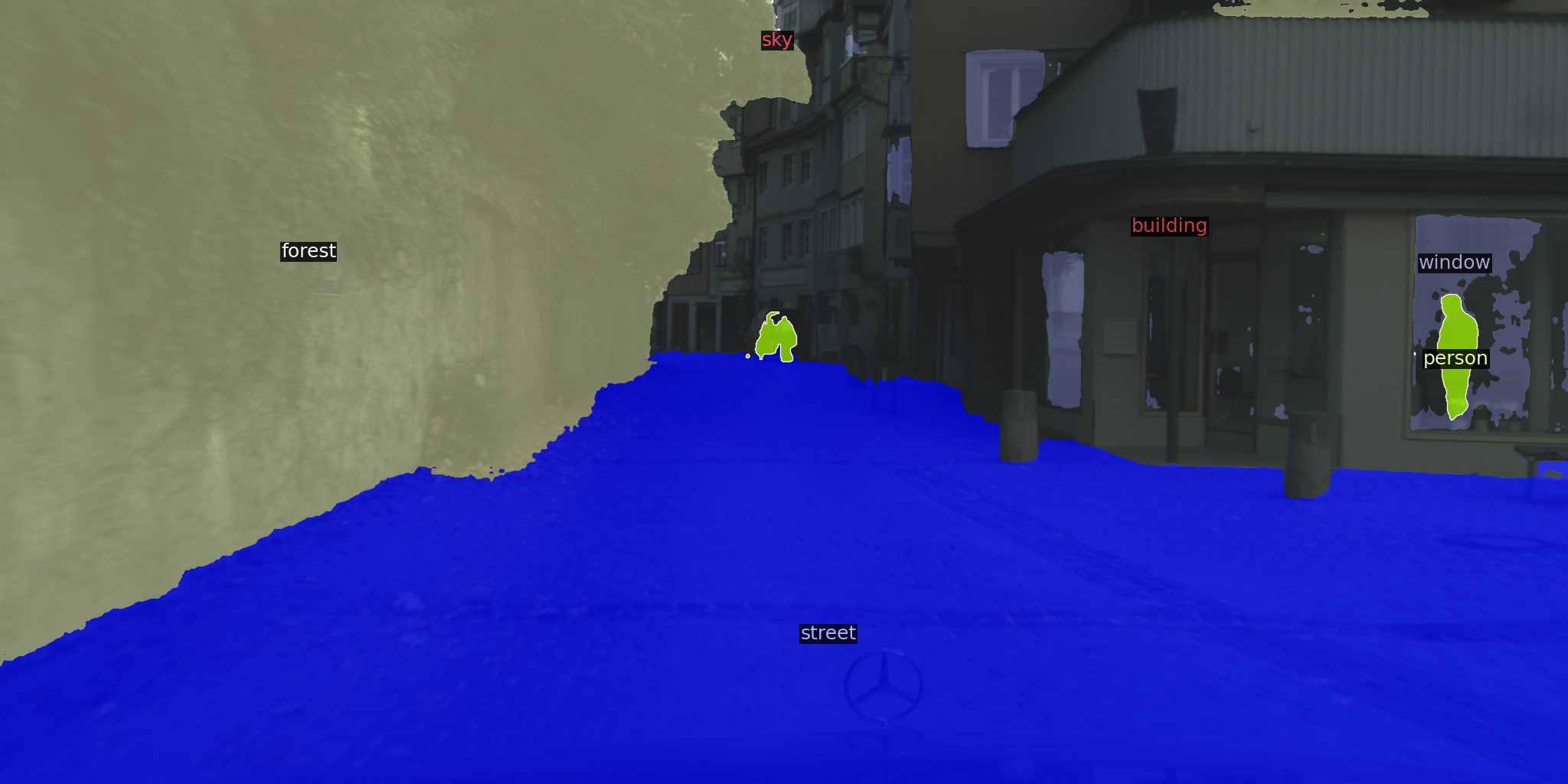}
  \caption{\textbf{Failure Case.} A wall with vegetation is mistaken for a forest, likely due to overclustering in the clustering process.}
\vspace{8pt}
\end{subfigure}
\caption{\textbf{Additional Qualitative Results} We discuss a selection of success cases and failure cases on Cityscapes above.}
\label{fig:qualitative_extra_city}
\end{figure}

In Fig.~\ref{fig:qualitative_extra_ade} and ~\ref{fig:qualitative_extra_city}, we provide additional qualitative results of \ours and discuss observed capabilities or limitations in the caption of each figure. Our main insights are that 1) \ours effectively labels unseen categories; 2) \ours uses semantically more precise classes for classification; 3) generated classes that are redundant could end up being segmented.


\newpage
\begin{small} 
\begin{algorithm}
\caption{LLM Mapper}
\begin{algorithmic}
\raggedright
\REQUIRE \textit{nouns, dataset\_vocabulary, llm\_batch\_size, prompt\_template,} \textbf{LLM}\vspace{5px}
\STATE \textit{all\_responses, map\_dict, skipped} $\leftarrow$ \textbf{initialize}()
\FOR{\textit{batch} in \textbf{split}(nouns, llm\_batch\_size)}
    \STATE \textit{dialogs} $\leftarrow$ \textbf{generateDialogs}(batch, prompt\_template)
    \STATE \textit{batch\_responses} $\leftarrow$ \textbf{LLM}(dialogs)
    \STATE \textbf{store}(all\_responses, (batch\_responses, batch))
\ENDFOR\vspace{5px}
\FOR{\textit{(batch\_response, batch)} in \textit{all\_responses}}
    \FOR{\textit{(response, noun)} in \textit{(batch\_response, batch)}}
        \STATE \textit{answer} $\leftarrow$ \textbf{parseResponse}(response)
        \STATE \textit{common} $\leftarrow$ \textbf{intersect}(answer, dataset\_vocabulary)
        \STATE \textbf{updateDict}(map\_dict, noun, common)
        \STATE \textbf{updateSkipped}(skipped, noun, common)
    \ENDFOR
\ENDFOR\vspace{5px}
\FOR{\textit{key} in \textit{map\_dict}}
    \IF{\textit{key} is in \textit{vocabulary}}
        \STATE \textit{map\_dict}[\textit{key}] $\leftarrow$ \textit{key}
    \ENDIF
    \IF{\textit{key} is in \textit{skipped}}
        \STATE \textit{skipped} $\leftarrow$ \textbf{removeItem}(\textit{skipped}, \textit{key})
    \ENDIF
\ENDFOR\vspace{5px}
\FOR{\textit{skipped\_key} in \textit{skipped}}
    \STATE \textit{map\_dict}[\textit{skipped\_key}] $\leftarrow$ ``background"
\ENDFOR\vspace{5px}
\RETURN \textit{map\_dict}
\end{algorithmic}
\label{algorithm}
\end{algorithm}
\end{small}

%% file: Tables/bboost.tex
\begin{table}[tb]
\caption{\textbf{BBoost Denoising and Caption Filtering Ablation.} Using both clustering and denoising for feature enhancement results in the highest mIoU performance on PASCAL VOC~\cite{VOC}, as well as the most compact vocabulary. We use 1 captioning cycle for simplicity.}
\centering
\footnotesize
\setlength{\tabcolsep}{3.5pt} 
\renewcommand{\arraystretch}{1.2}
\newcommand{\csp}{\hskip 2em}
\begin{tabular}{l ccc}
\toprule
\textbf{Enhancement Type} & \textbf{Caption Filtering} & \textbf{Vocabulary Size} & \textbf{mIoU} \\
\midrule
Clustering          & Disabled  & 4185 & \rd 3.8 \\
Clustering          & Enabled   & 1273 & \nd 70.4 \\
Clustering + Denoising    & Disabled  & 4026 & \rd 3.8 \\
Clustering + Denoising    & Enabled   & 938 & \fs 87.1  \\
\bottomrule
\end{tabular}
\label{tab:bboost}
\end{table}

%% file: Tables/mapper.tex
\begin{table}[tb]
\caption{\textbf{Ablation on Mappers.} Compared to other approaches for mapping, LAVE provides more reliable mappings between auto-vocabulary and target vocabulary classes for comparative evaluation that uses IoU. Tested on PASCAL VOC~\cite{VOC}.}
\centering
\footnotesize
\setlength{\tabcolsep}{36pt} 
\renewcommand{\arraystretch}{1.2}
\newcommand{\csp}{\hskip 2em}
\begin{tabular}{lc}
\toprule
\textbf{Mapper} & \textbf{mIoU}~\cite{VOC} \\
\midrule
Sentence-BERT~\cite{sentencebert} & \rd 73.9 \\
CLIP~\cite{clip}                  & \nd 77.5 \\
LAVE (Ours)                       & \fs 87.1 \\
\bottomrule
\end{tabular}
\label{tab:mapper}
\end{table}